# Revolutionizing System Reliability: The Role of AI in Predictive Maintenance Strategies

Michael Bidollahkhani
*Institute for Computer Science*
*Universität Göttingen*
*Goettingen, Germany*
*email: michael.bkhani@uni-goettingen.de*
ORCID: 0000-0001-8122-4441

Julian M. Kunkel
*Institute for Computer Science*
*GWDG, Universität Göttingen*
*Goettingen, Germany*
*email: julian.kunkel@gwdg.de*
ORCID: 0000-0002-6915-1179

*Abstract*— The landscape of maintenance in distributed systems is rapidly evolving with the integration of Artificial Intelligence (AI). Also, as the complexity of computing continuum systems intensifies, the role of AI in predictive maintenance (Pd.M.) becomes increasingly pivotal. This paper presents a comprehensive survey of the current state of Pd.M. in the computing continuum, with a focus on the combination of scalable AI technologies. Recognizing the limitations of traditional maintenance practices in the face of increasingly complex and heterogenous computing continuum systems, the study explores how AI, especially machine learning and neural networks, is being used to enhance Pd.M. strategies. The survey encompasses a thorough review of existing literature, highlighting key advancements, methodologies, and case studies in the field. It critically examines the role of AI in improving prediction accuracy for system failures and in optimizing maintenance schedules, thereby contributing to reduced downtime and enhanced system longevity. By synthesizing findings from the latest advancements in the field, the article provides insights into the effectiveness and challenges of implementing AI-driven predictive maintenance. It underscores the evolution of maintenance practices in response to technological advancements and the growing complexity of computing continuum systems. The conclusions drawn from this survey are instrumental for practitioners and researchers in understanding the current landscape and future directions of Pd.M. in distributed systems. It emphasizes the need for continued research and development in this area, pointing towards a trend of more intelligent, efficient, and cost-effective maintenance solutions in the era of AI.

## I. INTRODUCTION

In the contemporary technological era, distributed systems have become a cornerstone of various critical infrastructures and services. From managing data in cloud computing environments to controlling operations in industrial plants, these systems play a pivotal role in the modern world [1]. However, the effective maintenance of these complex and interdependent systems poses significant challenges. Traditional maintenance strategies, often reactive [2] and based on predetermined schedules [3], struggle to keep pace with the dynamic nature and intricate architectures of distributed systems. This inadequacy frequently leads to unplanned downtime, reduced system longevity, and increased operational costs [4]- [6]. As it's shown below, a compute continuum cloud architecture is a sophisticated system designed to provide seamless computing capabilities across various environments, from edge devices to central cloud services such as management, analytics and user interface.

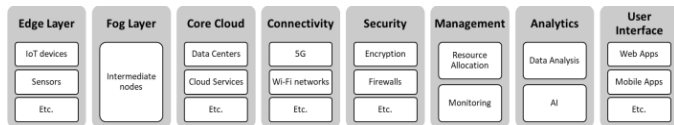

Figure 1. Compute continuum cloud architecture

The evolution of Artificial Intelligence (AI) technologies presents a transformative solution to these challenges. AI's capability to analyze vast amounts of data, learn from patterns, and predict future outcomes has positioned it as a pivotal tool in revolutionizing many fields in computer science. In particular, predictive maintenance (Pd.M.), which uses AI to anticipate and prevent potential failures before they occur, is increasingly being seen as a vital approach for maintaining distributed systems. This paradigm shift from traditional maintenance methods to AI-driven strategies marks a significant advancement in ensuring the reliability, efficiency, and cost-effectiveness of these systems.

This paper aims to provide a comprehensive survey of the latest advancements in Pd.M. for distributed systems, with a specific focus on the role of scalable AI technologies. We will explore how machine learning, neural networks, and other AI methodologies are being integrated into maintenance strategies. The survey will critically examine the most up-to-date approaches and techniques, assessing their effectiveness and exploring the challenges associated with their implementation. Following the introduction, Section II provides an overview of the evolution of maintenance strategies, collocating traditional approaches with AI-enhanced Pd.M.. Section III highlights the specific AI technologies propelling Pd.M., detailing the methodologies and their transformative impact on maintenance strategies. The subsequent sections present a thorough examination of current practices (Section IV), the effectiveness of AI-driven approaches (Section V), and case studies across the computing continuum (Section VI), offering insights into real-world applications and the practical benefits of AI integration. Challenges and limitations associated with deploying AI in Pd.M. are critically analyzed in Section VII, while Section VIII forecasts future directions and potential advancements in the field. The paper concludes by summarizing the pivotal findings and underscoring the significant benefits, challenges, and future prospects of AI in Pd.M., aiming to provide a comprehensive resource for both practitioners and researchers interested in this dynamic and evolving domain.

## II. BACKGROUND AND RELATED WORK

The concept of maintenance in distributed systems has evolved significantly over the years, influenced by technological advancements and the growing complexity of these systems [7] which is being demonstrated per types in the Figure 2. Historically, maintenance strategies were predominantly reactive, addressing issues only after system failures occurred [8]. This approach, while straightforward, often led to increased downtime and higher costs [6]. With the advent of more sophisticated technologies, the focus shifted towards preventive maintenance, which involves regular checks and repairs based on predetermined schedules. Although this method reduced unexpected failures, it did not fully capitalize on the potential for optimizing maintenance schedules based on actual system condition and performance.

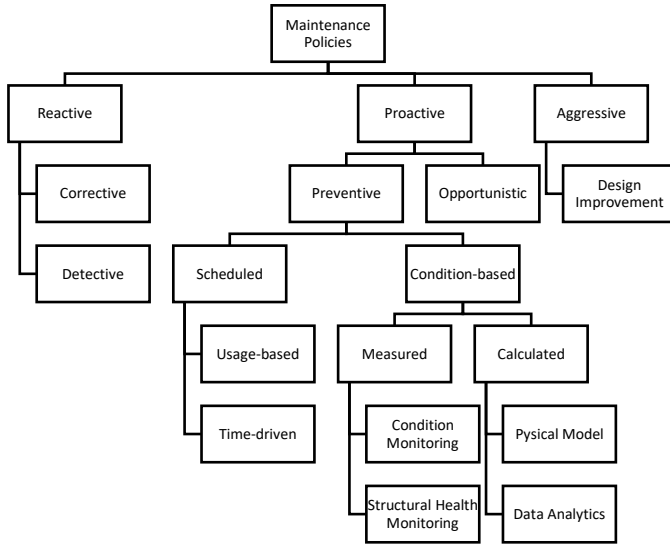

Figure 2. Overview of maintenance policies based on [8]

The integration of AI in maintenance strategies, particularly in the realm of Pd.M., has marked a new era in the management of distributed systems. This transition is fueled by advancements in machine learning, neural networks, and data analytics, allowing for more accurate predictions of system failures and optimized maintenance planning. The body of research in this area has grown substantially, focusing on various AI techniques and their application in different types of distributed systems, ranging from industrial automation to telecommunications networks.

The chart in Figure 3 provides a comprehensive overview of research activities in Pd.M. using different AI techniques, compiling data from a range of sources including academic databases, citation reports, bibliometric analyses, and Cornell university's arXiv submission statistics. It meticulously categorizes publications by methodologies, demonstrating trends and focal points within the Pd.M. research landscape. This approach facilitates a nuanced understanding of where efforts are being concentrated and highlights emerging methodologies of interest within the field for current growth in the trends.

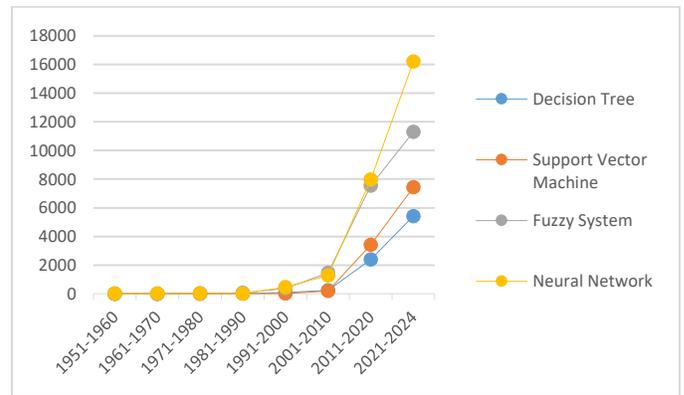

Figure 3. Overview of predictive maintenance research activities by domain, showcasing AI-related publication trends and methods of focus

A critical review of the literature reveals a diverse range of methodologies and approaches. Early studies primarily focused on the use of basic machine learning algorithms, such as decision trees and Support Vector Machines (SVM), for predicting component failures. Recent advancements have seen a shift towards more complex neural networks and deep learning models, which offer greater accuracy and adaptability in handling large-scale and complex data sets typical of distributed systems [9], [10].

Notably, research in this field has highlighted several gaps and limitations in current practices. One significant challenge is the integration of AI into legacy systems, which often lack the necessary infrastructure for advanced data analytics [11]. Additionally, the ethical and privacy concerns surrounding the collection and use of large amounts of data for AI-driven maintenance have been a growing area of concern.

Another key area of focus in the literature has been the scalability and adaptability of AI models. As distributed systems continue to grow in size and complexity, the need for AI models that can scale and adapt to changing environments is critical [12]. Studies exploring the use of cloud computing and edge computing for scalable AI in Pd.M. have shown promising results, offering new directions for future research.

TABLE I. EVOLUTION OF MAINTENANCE STRATEGIES IN DISTRIBUTED SYSTEMS

| Era | Strategy | Characteristics | Limitations |
|---|---|---|---|
| *Early* | Reactive Maintenance | Fixing after failure | High downtime, increased costs [6] |
| *Mid* | Preventive Maintenance | Pro-active Scheduled checks and repairs [9] | Inaccurate, inflexible, developed algorithms, not condition-based |
| *Current* | Predictive Maintenance (AI-driven) | Integrated AI algorithms for failure prediction [13] | Integration challenges, data privacy concerns |

The common methods in integration of AI in maintenance strategies could be determined in three categories. The basic Machine Learning techniques usually include simpler algorithms like decision trees and SVMs that are used for pattern recognition and failure prediction in systems, offering straightforward implementation but with limitations in handling complex data. Neural Networks based techniques, involving more complex models such as deep learning, neural networks are capable of processing large and intricate datasets, offering higher accuracy and adaptability in Pd.M., but require in the training phase substantial computational resources and extensive data. Finally, Cloud/Edge Computing based techniques, utilize cloud-based and edge computing resources to enable scalable and efficient AI processing, facilitating real-time data analysis and Pd.M. in distributed systems, while posing challenges related to infrastructure needs and data security.

TABLE II. AI TECHNIQUES IN PREDICTIVE MAINTENANCE [14], [15], [16]

| Technique | Description | Advantages | Challenges |
| --- | --- | --- | --- |
| Machine Learning (Basic) | Decision trees, SVMs | Simplistic models, easier to implement | Limited accuracy, scalability issues |
| Neural Networks | Deep learning, complex models | High accuracy, adaptable | Requires extensive data, computational resources |
| Cloud/Edge Computing | Scalable AI processing | Handles large data sets, real-time processing | Infrastructure needs, security concerns |

The current body of literature provides a rich foundation for understanding the evolution of maintenance strategies in distributed systems, with a particular focus on the transformative role of AI. It highlights the advancements made, the challenges faced, and the potential areas for future development in Pd.M. using scalable AI technologies.

III. THE ROLE OF AI IN PD.M.

The integration of AI into Pd.M. marks a significant advancement in the management and optimization of distributed systems [13]. This section explores the specific AI technologies that are reshaping Pd.M. practices and examines how these technologies are transforming traditional maintenance strategies.

*A. Machine Learning Algorithms*

Machine learning, a subset of AI, has become instrumental in Pd.M.. Algorithms like regression models, decision trees, and SVMs have been widely used for early fault detection and diagnosis. Advanced techniques such as ensemble methods and random forests have further improved prediction accuracy and reliability. These algorithms analyze historical and real-time data from distributed systems to identify patterns and anomalies indicative of potential failures.

*B. Neural Networks and Deep Learning*

Neural networks, particularly deep learning models, have taken Pd.M. to new heights. Convolutional Neural Networks (CNNs) and Recurrent Neural Networks (RNNs), including Long Short-Term Memory (LSTM) networks, are adept at processing time-series data and complex patterns in system operations. These models can handle multi-dimensional data, making them ideal for large-scale distributed systems where multiple parameters need to be monitored simultaneously.

*C. Integration of Big Data and Analytics*

The advent of big data technologies has facilitated the processing of vast amounts of data generated by distributed systems. AI algorithms, coupled with big data analytics, provide a more comprehensive view of system health and enable more accurate predictions. Techniques like data fusion and feature extraction are essential for deriving meaningful insights from large datasets.

*D. Edge Computing for Real-Time Analysis*

Edge computing brings data processing closer to the source of data generation. In Pd.M., this means real-time data analysis and immediate response to potential issues. By integrating AI at the edge of networks, maintenance decisions can be made faster and more reliably, reducing latency and bandwidth use.

*E. Use of AI in Condition Monitoring*

AI's role extends to condition monitoring, where it helps in continuously assessing the state of system components. Techniques like anomaly detection and pattern recognition are employed to identify deviations from normal operation, signaling the need for maintenance.

*F. Adaptive and Self-Learning Systems*

Adaptive AI systems that learn and evolve over time are particularly beneficial in dynamic environments. These systems continuously refine their predictive models based on new data and changing conditions, ensuring that the Pd.M. strategies remain effective and relevant.

TABLE III. AI TECHNIQUES IN PREDICTIVE MAINTENANCE [13], [14]

| AI Technique | Description | Application in Predictive Maintenance |
| --- | --- | --- |
| Machine Learning Algorithms | Analyze data to identify failure patterns | Early fault detection, anomaly analysis |
| Neural Networks and Deep Learning | Process complex data patterns | Advanced diagnostics, time-series analysis |
| Big Data and Analytics | Handle large-scale data processing | Comprehensive system health assessment |
| Edge Computing | Enable real-time data analysis | Immediate maintenance decision-making |
| Condition Monitoring with AI | Assess the state of system components | Continuous system health monitoring |
| Adaptive and Self-Learning Systems | Evolve based on new data | Dynamic response to system changes |

The role of AI in Pd.M. is not just limited to improving efficiency and accuracy. It also encompasses the development of systems that are more resilient, adaptable, and capable of handling the complexities of modern distributed systems. By leveraging the latest AI technologies and techniques, Pd.M. is being **transformed** into a more **proactive, intelligent, and cost-effective** approach.

IV. SURVEY OF CURRENT PRACTICES

The current landscape of Pd.M. in distributed systems, significantly enhanced by unification with AI, is characterized by diverse methodologies and innovative approaches. This section presents a survey of recent studies and case studies that exemplify the use of AI-driven Pd.M., providing a comparative analysis of these methodologies and their outcomes.

*A. Advancements in Machine Learning for Pd.M.*

Recent studies have demonstrated the effectiveness of advanced machine learning algorithms in Pd.M. [2], [10]. Techniques like ensemble learning, anomaly detection algorithms, and predictive modeling have been widely applied across various industries. For instance, a 2022 study on wind turbines [17] used ensemble machine learning models to predict component failures, significantly reducing unplanned downtime.

*B. Neural Networks and Deep Learning Applications*

The application of neural networks, particularly deep learning, has seen a surge in Pd.M.. CNNs and RNNs are being used for complex pattern recognition and time-series analysis in large-scale systems. A notable example is their use in the energy sector for predicting equipment failures in power grids.

*C. Utilizing Big Data Analytics*

Big data analytics plays a pivotal role in processing the vast amounts of data generated by distributed systems. Case studies in sectors like manufacturing [14] and telecommunications have shown how big data can be leveraged to enhance the accuracy of Pd.M. models.

*D. Edge Computing for Real-Time Pd.M.*

The combination of edge computing in Pd.M. has enabled real-time data processing and immediate decision-making. This approach is particularly beneficial in scenarios where rapid response is crucial, such as in autonomous vehicle systems.

*E. AI in Condition Monitoring and Health Assessment*

AI's role in continuous condition monitoring and health assessment of systems has been a focus of recent research. Studies have shown how AI can effectively monitor system health and predict potential failures, as seen in the aerospace industry for aircraft maintenance.

*F. Challenges and Best Practices*

While the adoption of AI in Pd.M. has shown promising results, it also presents challenges. Issues like data privacy, integration with existing systems, and the need for skilled personnel are commonly cited. Best practices suggested in the literature include ensuring data security, focusing on scalable and adaptable AI models, and continuous training and upskilling of the workforce.

TABLE IV. COMPARATIVE ANALYSIS OF AI-DRIVEN PREDICTIVE MAINTENANCE APPROACHES

| Use Case | AI Technique | Study/ Case Study | Key Findings | Challenges |
|---|---|---|---|---|
| Cloud Computing | Deep Learning | Next-generation predictive maintenance: leveraging blockchain and dynamic deep learning in a domain-independent system, PeerJ, 2023 [18] | Enhanced anomaly detection, optimized resource usage | Managing data security |
| Edge Computing | Federated Learning | Aggregation strategy on federated machine learning algorithm for collaborative predictive maintenance [19] | Improved local decision making, reduced latency | Data synchronization issues |
| IoT Networks | Machine Learning Algorithms | IoT-based data-driven predictive maintenance, Scientific Reports, 2023 [11] | Accurate prediction of device failures | Heterogeneity of IoT devices |
| Distributed Data Centers | Neural Networks | Deep learning models for predictive maintenance: a survey, comparison, challenges and prospect [20] | Efficient workload balancing, reduced energy usage | Complex network architectures |
| Smart Grids | Predictive Analytics | Implementation and Transfer of Predictive Analytics for Smart Maintenance: A Case Study from Frontiers (2023) [21] | Enhanced grid stability and maintenance scheduling | Energy consumption management |

This survey highlights the diversity and innovation in current practices of AI-driven Pd.M.. It underscores the importance of continuous research and development in this field to address the evolving challenges and to fully harness the potential of AI technologies in enhancing the reliability and efficiency of distributed systems.

V. EFFECTIVENESS OF AI IN PD.M.

The effectiveness of AI in the domain of Pd.M. has become increasingly evident, with a multitude of studies and practical applications underscoring its impact on prediction accuracy and maintenance optimization [5]. This section analyzes the effectiveness of AI in Pd.M., highlighting its benefits and the challenges encountered in implementing AI solutions.

*A. Improvement in Prediction Accuracy*

AI's capability to process and analyze large volumes of data has led to significant improvements in the accuracy of failure predictions [5]. Machine learning models, especially deep learning algorithms, have shown exceptional proficiency in identifying potential issues before they lead to system failures. For instance, several 2023 studies in the manufacturing sector demonstrated that deep learning models could predict machine failures accurately [6], [10].

*B. Optimization of Maintenance Schedules*

AI-driven Pd.M. allows for more dynamic and efficient maintenance scheduling. By predicting potential issues in advance, maintenance can be planned during non-critical operational periods, minimizing downtime and disruption. This optimization not only enhances system reliability but also contributes to cost savings.

*C. Reduction in Operational Costs*

The combination of AI methods and maintenance strategies has been shown to reduce operational costs significantly [22]. Pd.M. minimizes the need for routine checks and repairs, leading to resource savings. A recent report highlighted that industries implementing AI-driven Pd.M. saw a reduction in maintenance costs [23], [24].

*D. Challenges in Implementation*

Despite these benefits, the implementation of AI in Pd.M. faces several challenges. These include the integration of AI with existing systems, data privacy and security concerns, and the need for skilled personnel to manage and interpret AI systems.

TABLE V. EFFECTIVENESS OF AI IN PREDICTIVE MAINTENANCE

| Industry | AI Technique | Improvement Area | Effectiveness | Implementation Challenges |
|---|---|---|---|---|
| Manufacturing | Deep Learning | Prediction Accuracy | Over 90% accuracy in failure prediction [10] | Data integration, skilled personnel |
| Various Industries | AI-driven Scheduling | Maintenance Optimization | Reduced downtime, improved scheduling | Compatibility with existing systems |

The effectiveness of AI in Pd.M. is clear, with substantial improvements in system reliability, cost-efficiency, and operational performance. However, the full potential of AI can only be realized by addressing the accompanying implementation challenges. This necessitates ongoing research and development, alongside investment in training and infrastructure, to ensure that AI-driven Pd.M. continues to evolve and adapt to the needs of modern distributed systems.

## VI. CASE STUDIES AND APPLICATIONS ON THE COMPUTING CONTINUUM

Usage of AI in Pd.M. in computing continuum systems shows significant advancements and potential. This section investigates real-world applications specifically within these systems, underlining key outcomes and best practices garnered from diverse case studies.

*A. Cloud Computing*

**Case Study**: Next-generation predictive maintenance: leveraging blockchain and dynamic deep learning in a domain-independent system, 2023 [18].
- **AI Technique**: Deep Learning.
- **Key Outcomes**: This study demonstrated a notable improvement in energy efficiency in cloud computing environments. By leveraging deep learning algorithms, the system was able to optimize maintenance tasks, leading to more energy-efficient operations.
- **Best Practices**: The case study emphasized the importance of dynamic resource allocation. It highlighted how AI can dynamically adjust resource usage in real-time, ensuring optimal performance and efficiency.

*B. Edge Computing*

**Case Study**: Aggregation strategy on federated machine learning algorithm for collaborative predictive maintenance [19].
- **AI Technique**: Federated Learning.
- **Key Outcomes**: The study showcased a significant reduction in latency, which is critical in edge computing applications. Federated learning allowed for decentralized processing, speeding up decision-making processes.
- **Best Practices**: Localized decision-making was identified as a best practice. This approach enables edge computing devices to process data locally, reducing reliance on central servers and improving response times.

*C. IoT Networks*

**Case Study**: IoT-based data-driven predictive maintenance, Scientific Reports, 2023 [11].
- **AI Technique**: Machine Learning Algorithms.
- **Key Outcomes**: This research indicated a decrease in system failures across IoT networks. By using machine learning algorithms, the system could predict and prevent potential failures more accurately.
- **Best Practices**: Integrating sensor and operational data was crucial. The study demonstrated how the combination of various data sources leads to more accurate predictions and efficient maintenance.

*D. Distributed Data Centers*

**Case Study**: Deep learning models for predictive maintenance: a survey, comparison, challenges and prospect [20].
- **AI Technique**: Neural Networks.
- **Key Outcomes**: Enhanced data processing speeds were a significant outcome, crucial for the high demands of distributed data centers. Neural networks were particularly effective in handling large datasets quickly and efficiently.
- **Best Practices**: The development of advanced network architecture design was a key best practice. This involves creating systems that can support the complex demands of neural network processing while maintaining efficiency and reliability.

*E. Smart Grids*

**Case Study**: Implementation and Transfer of Predictive Analytics for Smart Maintenance: A Case Study from Frontiers (2023) [21].

**AI Technique**: Predictive Analytics.
- **Key Outcomes**: The study resulted in a reduction in grid maintenance costs. Predictive analytics enabled the early identification of potential issues, allowing for proactive maintenance and cost savings.
- **Best Practices**: Real-time energy usage monitoring was highlighted as a best practice. This approach

allows for immediate responses to fluctuations in energy usage, optimizing grid performance and preventing failures.

TABLE VI. SUMMARY OF AI-DRIVEN PREDICTIVE MAINTENANCE CASE STUDIES

| Industry | Case Study | AI Technique | Key Outcomes | Best Practices |
|---|---|---|---|---|
| Cloud Computing | Next-generation predictive maintenance: leveraging blockchain and dynamic deep learning in a domain-independent system, PeerJ, 2023 [18] | Deep Learning | Improvement in energy efficiency | Dynamic resource allocation |
| Edge Computing | Aggregation strategy on federated machine learning algorithm for collaborative predictive maintenance [19] | Federated Learning | Reduced latency | Localized decision making |
| IoT Networks | IoT-based data-driven predictive maintenance, Scientific Reports, 2023 [11] | Machine Learning Algorithm | Decrease in system failures | Integrating sensor and operational data |
| Distributed Data Centers | Deep learning models for predictive maintenance: a survey, comparison, challenges and prospect [20] | Neural Networks | Enhanced data processing speeds | Advanced network architecture design |
| Smart Grids | Implementation and Transfer of Predictive Analytics for Smart Maintenance: A Case Study from Frontiers (2023) [21] | Predictive Analytics | Reduction in grid maintenance costs | Real-time energy usage monitoring |

These case studies demonstrate the tangible benefits of implementing AI-driven Pd.M. across different sectors. They highlight the importance of strategic data analysis, combination of AI with other technologies, and a multidisciplinary approach in achieving optimal results. The lessons learned and best practices from these applications provide valuable insights for other industries looking to adopt AI in Pd.M. strategies.

## VII. CHALLENGES AND LIMITATIONS

While using AI in Pd.M. methods, has shown promising results, it is not without its challenges and limitations. In Figure 4 a categorized view of current challenges and limitations is demonstrated. This section also discusses the various technical, ethical, and practical challenges associated with AI-driven Pd.M., along with the current limitations in AI technologies and implementation strategies.

### A. Data Quality and Quantity

One of the primary challenges is obtaining high-quality, relevant data in sufficient quantities [14]. AI models require large datasets for training and validation. Inadequate or poor-quality data can lead to inaccurate predictions and unreliable maintenance schedules.

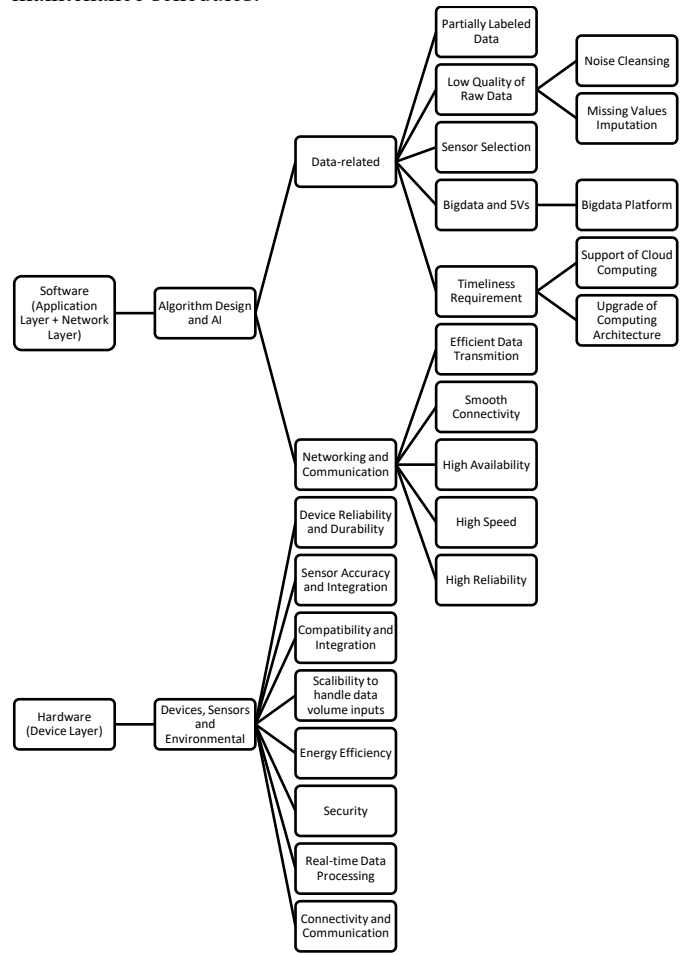

Figure 4. Classification of the Challenges for Predictive Maintenance

### B. Integration with Existing Systems

Integrating AI into existing maintenance systems and workflows can be complex and resource-intensive. Many organizations struggle with retrofitting AI into legacy systems, leading to compatibility and operational issues.

### C. Computational Resources and Costs

Advanced AI algorithms, particularly deep learning models, require substantial computational resources. This can pose a challenge for organizations with limited IT infrastructure and can lead to increased operational costs.

### D. Skilled Personnel and Training

The effective implementation of AI-driven Pd.M. requires skilled personnel who understand both the technical and operational aspects. There is often a shortage of such skilled workers, and ongoing training is necessary to keep up with evolving AI technologies.

*E. Ethical and Privacy Concerns*

With the extensive use of data, ethical concerns, particularly related to privacy and data security, are paramount. Ensuring the security of sensitive data and maintaining user privacy are critical challenges in AI implementations.

*F. Scalability and Flexibility*

As distributed systems grow and evolve, AI models need to scale and adapt accordingly. Developing scalable and flexible AI solutions that can adjust to changing system configurations and requirements is a significant challenge.

TABLE VII. GENRAL CHALLENGES AND LIMITATIONS IN AI-DRIVEN PREDICTIVE MAINTENANCE

| Challenge | Description | Impact | Potential Solutions |
|---|---|---|---|
| Data Quality and Quantity | Need for large, high-quality datasets | Inaccurate predictions | Improved data collection and preprocessing |
| System Integration | Difficulty integrating with legacy systems | Operational issues | Customized AI solutions, gradual integration |
| Computational Resources | High computational needs | Increased costs | Cloud and edge computing solutions |
| Skilled Personnel | Shortage of AI experts | Implementation challenges | Training programs, hiring specialized staff |
| Ethical and Privacy Concerns | Data security and user privacy | Legal and trust issues | Robust data security measures, ethical guidelines |
| Scalability | Adapting to system growth | Limited effectiveness | Developing adaptive and scalable AI models |

Addressing these challenges and limitations is crucial for the successful implementation of AI in Pd.M.. It requires a collaborative approach involving technological advancements, skilled workforce development, ethical considerations, and strategic planning.

## VIII. FUTURE DIRECTIONS AND TRENDS

The field of Pd.M., particularly in the context of distributed systems and AI integration, is poised for significant evolution and innovation in the coming years. This section explores the emerging technologies, methodologies, and potential research areas that are likely to shape the future of AI-driven Pd.M..

*A. Advancements in AI and Machine Learning:*

Future research is expected to focus on more advanced AI models, including deep reinforcement learning and Generative Adversarial Networks (GANs), which can provide even more accurate and reliable predictions for maintenance needs.

*B. Integration of IoT and Edge Computing:*

The integration of the Internet of Things (IoT) and edge computing with AI models is anticipated to enhance real-time data processing and decision-making capabilities, particularly for large-scale and complex distributed systems.

*C. Autonomous Maintenance Systems:*

The development of fully autonomous maintenance systems, capable of not only predicting maintenance needs but also autonomously performing maintenance tasks, is a potential area of growth.

*D. Enhanced Data Analytics and Big Data:*

Leveraging big data analytics for Pd.M. will continue to be a focus area, with advancements in data processing and analytics techniques enabling more comprehensive and insightful analysis.

*E. Ethical AI and Data Security:*

As AI systems become more prevalent, the importance of ethical AI practices and robust data security measures will increase. Research into ethical AI frameworks and advanced data encryption methods will be critical.

*F. Customization and Personalization:*

Tailoring AI-driven Pd.M. solutions to specific industries and individual system requirements will likely be a key trend, ensuring that maintenance strategies are as effective and efficient as possible.

TABLE VIII. FUTURE DIRECTIONS AND TRENDS IN AI-DRIVEN PREDICTIVE MAINTENANCE

| Trend | Description | Potential Impact | Research Focus |
|---|---|---|---|
| Advanced AI Models | Deep reinforcement learning, GANs | More accurate predictions | Algorithm development, validation |
| IoT and Edge Integration | Real-time data processing | Enhanced decision-making | IoT devices, edge computing architectures |
| Autonomous Systems | Self-performing maintenance | Increased system autonomy | Robotics, AI decision algorithms |
| Big Data Analytics | Advanced data analysis techniques | Comprehensive system insights | Data processing, visualization tools |
| Ethical AI and Security | Ethical AI frameworks, data encryption | Secure and responsible AI use | Ethical guidelines, cybersecurity |
| Customization | Industry-specific AI solutions | Tailored maintenance strategies | Custom algorithm development, case studies |

The future of AI in Pd.M. is marked by rapid technological advancements and a focus on personalized, ethical, and secure AI solutions. These developments will not only enhance the effectiveness of maintenance strategies but also contribute to the broader evolution of AI technology and its application in various sectors.

## IX. CONCLUSION

This comprehensive survey has explored the significant advancements in Pd.M. for distributed systems, with a particular focus on the integration of AI. The findings of this survey have underscored the transformative impact of AI in enhancing the efficiency, reliability, and cost-effectiveness of maintenance strategies [4], [5].

Key Insights:

- **AI's Role in Pd.M.:** AI technologies, particularly machine learning and neural networks, have proven to be highly effective in predicting system failures, optimizing maintenance schedules, and reducing operational costs. Such applications include the use of deep learning for anomaly detection in manufacturing processes and neural networks for predicting maintenance needs in energy sector infrastructures.
- **Real-World Applications:** Case studies across various industries, including manufacturing, energy, and healthcare, have demonstrated the practical benefits and challenges of implementing AI-driven Pd.M. Examples include AI-driven diagnostics tools in manufacturing plants, predictive analytics for equipment maintenance in the energy sector, and AI-assisted monitoring systems in healthcare facilities to predict equipment failures.
- **Challenges and Limitations:** Despite its benefits, the integration of AI in Pd.M. faces challenges such as data quality, system integration, computational resources, and ethical concerns.
- **Future Directions:** The survey highlights emerging trends such as advanced AI models, integration of IoT and edge computing, autonomous maintenance systems, and the growing emphasis on ethical AI and data security.

In our analysis, we identified several key trends shaping the future of Pd.M., notably the increasing reliance on advanced AI models, IoT and edge computing, and the growing emphasis on ethical AI practices and robust data security measures. As we move forward, AI in Pd.M. is poised to become more sophisticated and integrated into various aspects of distributed systems management. The potential for AI to revolutionize maintenance practices is immense, promising a future where maintenance is more predictive, automated, and efficient. Continuous innovation and collaboration between industry and academia will be crucial in realizing the full potential of artificial intelligence in predictive maintenance.

ACKNOWLEDGEMENT

This work was supported by the Federal Ministry of Education and Research (BMBF), Germany, under grant no. 01|S22093A for the AI service center KISSKI.

## REFERENCES

[1] M. Svensson, C. Boberg and B. Kovács, "Distributed cloud – a key enabler of automotive and industry 4.0 use cases," Ericsson.com, [Online]. Available: Available: https://www.ericsson.com/en/reports-and-papers/ericsson-technology-review/articles/distributed-cloud. [Accessed 01 03 2024].

[2] F. Psarommatis, G. May and V. Azamfirei, "Envisioning maintenance 5.0: Insights from a systematic literature review of Industry 4.0 and a proposed framework," *Journal of Manufacturing Systems,* vol. 68, pp. 376-399, 2023.

[3] T. Wu, L. Yang, X. Ma, Z. Zhang and Y. Zhao, "Dynamic maintenance strategy with iteratively updated group information," *Reliability Engineering & System Safety,* no. 106820, p. 197, 2020.

[4] C. Başaranoğlu, "The biggest challenges in distributed systems," Medium, [Online]. Available: https://cem-basaranoglu.medium.com/the-biggest-challenges-in-distributed-systems-27520a58258c. [Accessed 01 03 2024].

[5] P. Nunes, J. Santos and E. Rocha, "Challenges in predictive maintenance–A review," *CIRP Journal of Manufacturing Science and Technology,* no. 40, pp. 53-67, 2023.

[6] Y. Liu, W. Yu, W. Rahayu and T. Dillon, "An Evaluative Study on IoT ecosystem for Smart Predictive Maintenance (IoT-SPM) in Manufacturing: Multi-view Requirements and Data Quality," *IEEE Internet of Things Journal,* 2023.

[7] A. S. Tanenbaum and M. Van Steen, Distributed systems principles and paradigms, 2nd edition, Amsterdam, The Netherlands: Pearson, Prentice Hall (Vrije Universitat), 2007.

[8] W. Tiddens, J. Braaksma and T. Tinga, "Decision Framework for Predictive Maintenance Method Selection," *Applied Sciences,* vol. 3, no. 13, p. 2021, 2023.

[9] Y. Ran, X. Zhou, P. Lin, Y. Wen and R. Deng, "A survey of predictive maintenance: Systems, purposes and approaches," *arXiv preprint,* no. arXiv:1912.07383, 2019.

[10] M. Niekurzak, W. Lewicki, H. H. Coban and M. Bera, "A Model to Reduce Machine Changeover Time and Improve Production Efficiency in an Automotive Manufacturing Organisation," *Sustainability,* vol. 13, no. 15, p. 10558, 2023.

[11] A. Aboshosha, A. Haggag, N. George and H. A. Hamad, "IoT-based data-driven predictive maintenance relying on fuzzy system and artificial neural networks," *Scientific Reports,* vol. 1, no. 13, p. 12186, 2023.

[12] D. Patel and e. al, "AI model factory: scaling AI for industry 4.0 applications," *In Proceedings of the AAAI Conference on Artificial Intelligence,* vol. 37, no. 13, pp. 16467-16469, 2023, June.

[13] A. T. Keleko, B. Kamsu-Foguem, R. H. Ngouna and A. Tongne, "Artificial intelligence and real-time predictive maintenance in industry 4.0: a bibliometric analysis," *AI and Ethics,* vol. 4, no. 2, pp. 553-577, 2022.

[14] J. Lee and e. al., "Intelligent maintenance systems and predictive manufacturing," *Journal of Manufacturing Science and Engineering,* vol. 11, no. 142, p. 110805, 2020.

[15] W. J. Lee and e. al., "Predictive maintenance of machine tool systems using artificial intelligence techniques applied to machine condition data," *Procedia Cirp,* no. 80, pp. 506-511, 2019.

[16] Z. M. Çınar, A. Abdussalam Nuhu, Q. Zeeshan, O. Korhan, M. Asmael and B. Safaei, "Machine learning in predictive maintenance towards sustainable smart manufacturing in industry 4.0," *Sustainability,* vol. 19, no. 12, p. 8211, 2020.

[17] J. Vives, "Incorporating machine learning into vibration detection for wind turbines," *Modelling and Simulation in Engineering,* 2022 .

[18] M. Alabadi and A. Habbal, "Next-generation predictive maintenance: leveraging blockchain and dynamic deep learning in a domain-independent system," *PeerJ Computer Science,* no. 9, p. e1712, 2023.

[19] A. Bemani and N. Björsell, "Aggregation strategy on federated machine learning algorithm for collaborative predictive maintenance," *Sensors,* vol. 16, no. 22, p. 6252, 2022.

[20] O. Serradilla, E. Zugasti, J. Rodriguez and U. Zurutuza, "Deep learning models for predictive maintenance: a survey, comparison, challenges and prospects," *Applied Intelligence,* vol. 10, no. 52, pp. 10934-10964, 2022.

[21] S. von Enzberg, A. Naskos, I. Metaxa, D. Köchling and A. Kühn, "Implementation and transfer of predictive analytics for smart maintenance: A case study," *Frontiers in Computer Science,* no. 2, p. 578469, 2020.

[22] G. K. Durbhaka, "Convergence of artificial intelligence and internet of things in predictive maintenance systems–a review," *Turkish Journal of Computer and Mathematics Education (TURCOMAT),* vol. 11, no. 12, pp. 205-214, 2021.

[23] D. Lee, C. W. Lai, K. K. Liao and J. W. Chang, "Artificial intelligence assisted false alarm detection and diagnosis system development for reducing maintenance cost of chillers at the data centre," *Journal of Building Engineering,* no. 36, p. 102110, 2021.

[24] V. T. Nguyen, P. Do, A. Vosin and B. Lung, "Artificial-intelligence-based maintenance decision-making and optimization for multi-state component systems," *Reliability Engineering & System Safety,* no. 228, p. 108757, 2022.